\documentclass[lettersize,journal]{IEEEtran}
\usepackage{amsmath,amsfonts}
\usepackage{algorithmic}
\usepackage{algorithm}
\usepackage{array}
\usepackage[caption=false,font=normalsize,labelfont=sf,textfont=sf]{subfig}
\usepackage{textcomp}
\usepackage{stfloats}
\usepackage{url}
\usepackage{verbatim}
\usepackage{graphicx}
\usepackage{cite}
\usepackage{multirow}
\newcommand{\etal}{\textit{et al.}}
\begin{document}

\title{3D Reconstruction from Transient Measurements with Time-Resolved Transformer}
\author{Yue Li, Shida Sun, Yu Hong, Feihu Xu, and Zhiwei Xiong
		\thanks{The authors are with University of Science and Technology of China, Hefei 230026, China (E-mail: yueli65@mail.ustc.edu.cn, sdsun@mail.ustc.edu.cn, hy520484@mail.ustc.edu.cn, 
        feihuxu@ustc.edu.cn, zwxiong@ustc.edu.cn).
        }}
\markboth{Journal of \LaTeX\ Class Files,~Vol.~14, No.~8, August~2021}%
{Shell \MakeLowercase{\textit{et al.}}: A Sample Article Using IEEEtran.cls for IEEE Journals}

\IEEEpubid{0000--0000/00\$00.00~\copyright~2021 IEEE}
\maketitle

\begin{abstract}

Transient measurements, captured by the time-resolved systems, are widely employed in photon-efficient reconstruction tasks, including line-of-sight (LOS) and non-line-of-sight (NLOS) imaging. However, challenges persist in their 3D reconstruction due to the low quantum efficiency of sensors and the high noise levels, particularly for long-range or complex scenes. To boost the 3D reconstruction performance in photon-efficient imaging, we propose a generic Time-Resolved Transformer (TRT) architecture. Different from existing transformers designed for high-dimensional data, TRT has two elaborate attention designs tailored for the spatio-temporal transient measurements. Specifically, the spatio-temporal self-attention encoders explore both local and global correlations within transient data by splitting or downsampling input features into different scales. Then, the spatio-temporal cross attention decoders integrate the local and global features in the token space, resulting in deep features with high representation capabilities. Building on TRT, we develop two task-specific embodiments: TRT-LOS for LOS imaging and TRT-NLOS for NLOS imaging. Extensive experiments demonstrate that both embodiments significantly outperform existing methods on synthetic data and real-world data captured by different imaging systems. In addition, we contribute a large-scale, high-resolution synthetic LOS dataset with various noise levels and capture a set of real-world NLOS measurements using a custom-built imaging system, enhancing the data diversity in this field. Code and datasets are available at \url{https://github.com/Depth2World/TRT}.

\end{abstract}


\section{Introduction}

The emergence of time-solved sensors, e.g., single-photon avalanche diode (SPAD), with  photon detection capability and picosecond-level time resolution, 
has significantly revolutionized active 3D sensing. These innovations facilitate the accurate detection of extremely weak photon signals, even from ultra-long distances or attenuated, diffusely scattered paths. As a result, time-solved systems have been proven highly effective in a range of applications, including lifetime imaging microscopy~\cite{gyongy2019fluorescence,bruschini2019single,hirvonen2020fast}, line-of-sight imaging~\cite{kirmani2014first,li2020single,po2022adaptive,lee2023caspi}, and non-line-of-sight imaging~\cite{velten2012recovering,lindell2019wave,liu2019non}.
Such imaging systems typically consist of a laser source, a SPAD, and a time-correlated single-photon counting (TCSPC) sensor. The pulsed laser provides the system synchronization signal and works as the light source, emitting periodic light directed towards the scenarios in line-of-sight or out of line-of-sight through the relay wall. The SPAD and TCSPC capture the returning photons and record their time-of-arrival during each pulse cycle. The system accumulates the photon data over multiple pulse cycles at each scanning point, forming the histogram. By collecting data across all designated scanning points, the spatio-temporal transient measurements are captured, and the 3D spatial volume can be reconstructed by modeling the travel of light.

Existing 3D reconstruction algorithms for photon-efficient transient measurements ~\cite{shin2016photon,rapp2017few,lee2023caspi,o2018confocal,heide2019non,lindell2019wave,liu2019non} have achieved decent results, but are still confronted with great challenges. 
A primary issue is the inherently low signal-to-background ratio (SBR) in transient measurements. Due to the low quantum efficiency of the photon detector and the high noise levels caused by the ambient light and the dark count, these measurements are often dominated by noise, with only a limited number of effective photons, making high-quality reconstruction difficult. For example, in line-of-sight single-photon imaging, it is common to detect fewer than one photon per pixel on average~\cite{li2020single,li2021single}, leading to texture loss and increased background noise. In non-line-of-sight imaging, the measurements contain substantial noise but few effective photons, leading to strong assumptions for reconstruction ~\cite{heide2014diffuse,o2018confocal,heide2019non}, or high sensitivity to depth variations ~\cite{lindell2019wave,liu2019non}
. 
On the other hand, the spatio-temporal transient measurements are inherently high-dimensional and sparse. Although deep neural networks~\cite{lindell2018single,chen2020learned,chopite2020deep} have made progress in terms of reconstruction performance, existing methods often overlook the intrinsic characteristics of the transient measurements and fail to adequately exploit the spatio-temporal correlation. 
Consequently, there still remains a large room for boosting their performance in complex scenes and generalization capabilities across diverse real-world systems. 

\IEEEpubidadjcol

Transformer~\cite{vaswani2017attention}, originally designed for natural language processing, has become a foundational architecture across various domains~\cite{parmar2018image,han2021transformer,liu2021swin} due to its ability to model long-range dependencies in data. Recent transformer variants extend this capability to high-dimensional visual data, e.g., video sequences~\cite{yang2022tubedetr,hsu2023video}, 3D voxels~\cite{asadi2023transformer,lyu2024iterative}, and point clouds~\cite{yao2022dynamic,fan2022point,wei2022spatial}. 
However, applying transformers to spatio-temporal transient measurements meets challenges. First, directly flattening the spatio-temporal dimensions leads to extremely long token sequences, making standard self-attention computationally intractable. Second, this flattening process disrupts the inherent structure of transient data by mixing spatial and temporal axes, particularly the spatio-temporal correlation imposed by light transport.



In this work, we propose a generic Time-Resolved Transformer (TRT) architecture for 3D reconstruction from transient measurements, which leverages the powerful representation capability of transformer for capturing
local and global spatio-temporal correlations in transient measurements. Specifically, to exploit these correlations, we elaborate two  attention designs, i.e., spatio-temporal self-attention (STSA) and spatio-temporal cross attention (STCA). Given the extracted shallow features from the transient measurements, two STSA encoders are employed to extract local and global information from the shallow features, respectively. For the local encoder, the input features are split into patches, and the local information is exploited in each patch along the spatial and temporal dimensions, successively. For the global encoder, the input features are downsampled to a smaller scale, and the global information is exploited along spatial and temporal dimensions in the whole feature space. The complementary local and global information is further integrated with each other into the token space of transformers by the STCA decoders, generating deep local and global features with high representation capabilities. Finally, the deep local and global features are fused together for the subsequent modules which reconstruct the 3D volume of scenes.

We adapt the proposed TRT to two representative 3D reconstruction tasks from transient measurements, including line-of-sight (LOS) single-photon imaging and non-line-of-sight (NLOS) imaging. The embodiments are termed as TRT-LOS and TRT-NLOS, respectively. TRT-LOS is applied to reconstruct 3D scenes from direct photon-efficient measurements acquired by time-resolved sensors. The transient inputs are encoded, processed through TRT blocks, and fused to recover the spatial structure of the target objects. TRT-NLOS is applied to the more challenging setting, where the reconstruction involves occluded scenes inferred from indirect light transport. Different from TRT-LOS, a lightweight denoising head is designed to enhance the input data quality. TRT then extracts and fuses spatial-temporal features informed by physical priors, ultimately reconstructing the hidden geometry and appearance. Extensive experiments are performed on both synthetic and real-world datasets. Compared with existing traditional and deep-learning-based solutions, TRT-LOS and TRT-NLOS achieve superior reconstruction performance as well as improved generalization capability to real-world scenarios. In addition to the publicly available data, we introduce a large-scale, high-resolution synthetic LOS dataset for model training and a complex dataset for test. Also, we capture a set of real-world NLOS measurements with a self-built imaging system. 

A preliminary version of TRT was presented in~\cite{li2023nlost}, namely NLOST, where it was specifically designed for NLOS imaging. In this work, we further promote a generic transformer architecture to achieve high-performance 3D reconstruction from transient measurements by making the following extensions. 1) We generalize TRT to LOS imaging, demonstrating the versatility and general applicability of the proposed architecture. 2) We design a transient measurement denoiser for TRT-NLOS, which substantially improves the quality of the input data and enhances the performance of subsequent reconstruction. 3) We offer an expanded literature review, clearer motivation behind TRT, more extensive experimental settings and results, broader application scenarios, and in-depth discussions.

  The contributions of this work are summarized as follows:
\begin{itemize}
\item A generic time-resolved transformer (TRT) architecture, incorporating two elaborate attention designs, is proposed to exploit the complementary local and global correlations within the spatio-temporal transient measurements.
\item TRT has been successfully applied to two representative 3D reconstruction tasks: line-of-sight (LOS) imaging and non-line-of-sight (NLOS) imaging. The proposed methods achieve state-of-the-art performance on both synthetic and real-world data, while exhibiting robust generalization capabilities across different imaging systems.
\item A large-scale, high-resolution synthetic LOS training dataset with various noise levels and a complex test set are introduced. Besides, a set of real-world NLOS measurements is captured with a self-built confocal system. Both datasets have been released to facilitate future advancements in this field.
\end{itemize}

\section{Related Work}

\subsection{Time-Resolved Photon Detection}
Advancements in optics have enabled the detection of individual photons, leveraging technologies such as single-photon avalanche diode (SPAD)~\cite{hadfield2009single,richardson2009low}, jot-based CMOS sensors~\cite{fossum2005sub,ma2017photon}, and Geiger-mode avalanche photodiode (APD). 
These time-resolved sensors have been employed in novel imaging systems with unique statistical characteristics, including TCSPC~\cite{pediredla2018signal}, inter-photon timing~\cite{gupta2019photon}, and free-running configurations~\cite{ingle2019high, wei2023passive}. In this paper, we investigate SPAD-based light detection and ranging (LiDAR) systems that integrate TCSPC devices to achieve 3D perception through the time-of-flight (ToF) principle. Compared to commercial LiDAR systems based on APDs, SPAD-based systems demonstrate superior photon sensitivity and temporal resolution. These attributes make them highly robust for applications such as long-range LOS imaging, and even NLOS imaging by detecting photons reflected from occluded objects.

\subsection{Line-of-Sight Single-Photon Imaging}
The active single-photon imaging systems employ a pulsed laser source which emits periodic pulses towards the targeted scenarios, and the detector SPAD collects the reflected photons with temporal stamps. The spatio-temporal transient measurement contains the histograms of pre-set scanning points, and the depth map can be derived from the raw photons. During the capturing process, only a few effective signals arrive at the system, while ambient light and noise degrade the transient measurement, leading to reconstruction challenges. 

\noindent \textbf{Traditional LOS Reconstruction.}  Many computational reconstruction algorithms~\cite{mccarthy2013kilometer,kirmani2014first,shin2015photon,shin2016photon,rapp2017few,li2020single,lee2023caspi} have been proposed for single-photon imaging. Kirmani~\etal~\cite{kirmani2014first} focus on the first photon and leverage the spatial correlations to enhance the reconstruction from low-flux scenes. Shin~\etal~\cite{shin2015photon,shin2016photon} introduce a physically accurate forward model within the photon counting statistics and spatial correlations, and they develop a robust method for reconstructing the depth and reflectivity. Rapp~\etal~\cite{rapp2017few}
further enhance the low-light reconstruction performance
by unmixing the contribution of signals and noise.
Li~\etal~\cite{li2020single,li2020super,li2021single} take the multiple-return issue into long-ranging tasks and devise iterative methods for 3D imaging. Lee~\etal~\cite{lee2023caspi}
exploit the local and non-local correlations and achieve estimating the scenes under extremely challenging conditions. However, traditional methods rely on strict assumptions and pre-defined parameters, which hinder their real-world applications to different imaging systems.

\noindent \textbf{Deep LOS Reconstruction.}  Lindell~\etal~\cite{lindell2018single} propose the first deep learning approach to photon-efficient 3D imaging, which fuses the input intensity image and estimates the depth map. Peng~\etal~\cite{peng2020photon,peng2022boosting} analyze the long-range correlations and introduce a non-local neural network, which achieves decent reconstruction fidelity. 
Tan~\etal~\cite{tan2020deep} solve the multiple return issue in long-range imaging as a deblurring task. Yao~\etal~\cite{yao2022dynamic} employ the sparse convolution for inference acceleration.
These data-driven approaches have shown adaptability across different imaging systems and improved reconstruction performance. However, 
they still overlook the exploitation of spatio-temporal features within transient measurements, leaving the potential for achieving higher reconstruction performance untapped.

\subsection{Non-Line-of-Sight Imaging}

In active NLOS imaging systems ~\cite{kirmani2009looking,velten2012recovering,heide2014diffuse,o2018confocal,lindell2019wave,liu2019non},
a laser source projects a short-pulse light to the relay wall. The light propagates from the relay wall to the hidden object, then reflects back to the relay wall and is finally captured with a time-resolved SPAD detector. The hidden volume could be reconstructed by modeling the three bounces of the traveling light, achieving ``seeing around corners".

\noindent \textbf{Traditional NLOS Reconstruction.}
Many algorithms have been developed for time-resolved NLOS reconstruction since Kirmani~\etal~\cite{kirmani2009looking} propose to recover the hidden object out of the visible line of sight. As a precursory work in this field, Velten~\etal~\cite{velten2012recovering} propose a filtered back-projection (FBP) method to recover the hidden objects from NLOS measurements. 
O'Toole~\etal~\cite{o2018confocal} facilitate the light-cone transform (LCT) for NLOS reconstruction under the following assumptions: light scatters isotropically and only once behind the wall, and the scene contains no occlusions. They simplify the transient formation in a linear 3D convolution form, and the reconstruction can be expressed as a deconvolution process and solved efficiently. Following~\cite{o2018confocal}, Heide~\etal~\cite{heide2019non} further model the partial occlusions and surface normals in NLOS imaging and develop a factorization approach for nonlinear inverse time-resolved light transport. 
Recent researches have transitioned from geometrical optics models to wave propagation models~\cite{lindell2019wave,liu2020phasor}.
Lindell~\etal~\cite{lindell2019wave} introduce a wave-based image formation model for NLOS imaging and adopt frequency-wavenumber migration (FK).
Liu~\etal~\cite{liu2020phasor} start from the phasor field formalism and present a Rayleigh Sommerfeld Diffraction (RSD) algorithm for non-confocal data. However, the traditional algorithms are either restricted by the ideal assumptions or fragile for distant targets in real-world scenarios.

\noindent \textbf{Deep NLOS Reconstruction.}
Chopite~\etal~\cite{chopite2020deep} first employ a convolutional neural network for NLOS depth estimation, with a 3D encoder and a 2D decoder in the U-Net~\cite{cciccek20163d} architecture. Due to the lack of special network design for the transient measurement, this model behaves no better than physics-based solutions~\cite{o2018confocal,lindell2019wave,liu2020phasor} 
Chen~\etal~\cite{chen2020learned} propose a learned feature embedded network (LFE) to reduce the domain gap between synthetic and real-world data, which incorporates the physics-based method~\cite{lindell2019wave} at the feature level and then projects the features from 3D spatial domains to 2D planes directly to reconstruct final intensity and depth maps. While promising results are achieved, the 3D to 2D projection may lead to information loss, and LFE requires multi-view supervision during training which burdens the training data generation. Inspired by the recently proposed Neural Radiance Field (NeRF), Shen~\etal~\cite{shen2021non} introduce Neural Transient Field (NeTF) to recover the 3D volume from transient measurements, which uses the multi-layer perception to represent a 3D density volume. Nevertheless, NeTF suffers from severe noise on smooth surfaces when recovering the geometry. In addition, the transient field has to be rendered for each measurement, which poses a huge computational burden during inference. Yu~\etal~\cite{yu2023enhancing} propose a learnable inverse kernel to alleviate the spectral basis. By manually differentiating the high and low frequency parts, this method enhances the detail reconstruction. However, the model learns the point spread function of the imaging system but is trained on synthetic data, hindering its generalization on the real-world measurements. Overall, there is still a large room for performance improvement in terms of NLOS reconstruction.

\subsection{Transformer for High Dimensional Data}
Transformer has been widely used in vision tasks~\cite{jaderberg2015spatial,vaswani2017attention,parmar2018image,yang2020learning,girdhar2019video,arnab2021vivit,li2021revisiting,han2021transformer,liu2021swin}. Recently, researchers apply transformers to 3D reconstruction tasks, such as depth estimation~\cite{li2021revisiting,su2022chitransformer} and point cloud registration~\cite{yew2022regtr}. To name a few, Li~\etal~\cite{li2021revisiting} make use of multiple layer of self and cross attention for stereo matching. Su~\etal~\cite{su2022chitransformer} further leverage the attention mechanism to rectify cues for depth estimation. Yew ~\cite{yew2022regtr} propose the cross attention encoder to replace the explicit feature matching in point cloud registration. On the other hand, transformer variants are also applied to high-dimensional data, such as video sequences~\cite{yang2022tubedetr,hsu2023video}, 3D voxels~\cite{asadi2023transformer,lyu2024iterative}, and point clouds~\cite{yao2022dynamic,fan2022point,wei2022spatial}. 
While these models capture the relationships between 3D elements and facilitate rich interactions between features, they are not directly applicable to photon-efficient measurements that exhibit totally different characteristics.
In this paper, we propose a generic transformer architecture for 3D reconstruction from transient measurements, which leverages special attention designs for capturing the short-range and long-range spatio-temporal correlations in the transient measurements.

\begin{figure*}[!t]
	\centering
	\includegraphics[width=\textwidth]{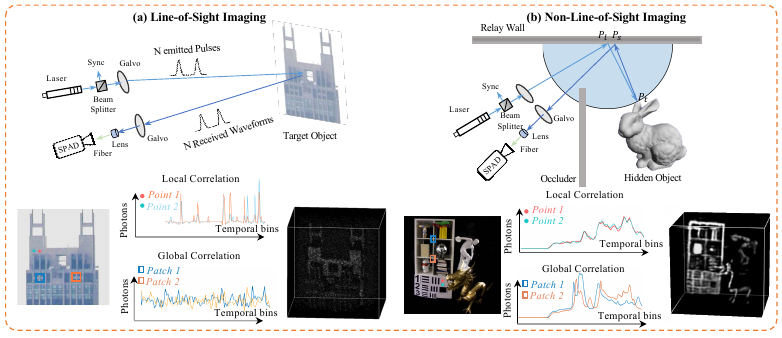}
	\caption{The schematic diagrams of line-of-sight imaging and non-line-of-sight imaging, with the example of the local and global correlations in the transient measurements. In line-of-sight imaging, the scene is oriented towards the system, while the scene faces the relay wall in non-line-of-sight imaging. The points and patches shown in the diagram refer to the regions within the scanning area that correspond to the orientation of the scene.}
\label{fig:correlations}
\end{figure*}


\section{Time-Resolved Transformer}
In this section, we first delineate the local and global correlations within the transient measurement. We then introduce the time-resolved transformer, which comprises spatio-temporal self-attention encoder that captures both local and global features at a deep level, and spatio-temporal cross attention decoders that integrate these deep features.
\subsection{Correlations in Transient Measurements}
\noindent \textbf{Local Correlation.}
For natural scenes, a certain location usually has similar intensity and depth values to its neighborhoods. This short-range correlation (denoted as local correlation) generally holds in a small range of regions and has been exploited in many vision tasks, e.g., image segmentation~\cite{alush2015hierarchical,li2021contexts}, edge detection~\cite{canny1986computational,jing2022recent}, denoising~\cite{hu2021pseudo,ilesanmi2021methods}, etc. For example, in the LOS/NLOS measurements as shown in Fig.~\ref{fig:correlations}, we can find that the histograms of two adjacent points in the transient data are close to each other, which demonstrates that the spatio-temporal measurements also contain local correlation. This kind of correlation is exploited by our proposed spatio-temporal self-attention encoder under the constraint of local continuity for the reconstructed scene.

\noindent \textbf{Global Correlation.}
For natural scenes, distant patches with similar geometry may have similar intensity and depth values. This long-range correlation (denoted as a global or nonlocal correlation) generally exists in different regions of the scene, which has also been exploited in many vision tasks, e.g., image matching~\cite{chen2022aspanformer,xu2022gmflow}, inpainting~\cite{zheng2021nonlocal,quan2022image}, restoration~\cite{liu2018non,mou2021dynamic}, etc. For the LOS/NLOS measurements as shown in Fig.~\ref{fig:correlations}, if we average the histograms of two patches with similar geometry in the transient data, the resulting curves are quite similar. It suggests that the global correlation also holds in the spatio-temporal measurements for LOS and NLOS imaging. This kind of correlation is further exploited by our proposed spatio-temporal self-attention encoder under the constraint of global consistency for the reconstructed scene.

\subsection{Attention Designs}

\begin{figure}[!t]
	\centering
	\includegraphics[width=0.49\textwidth]{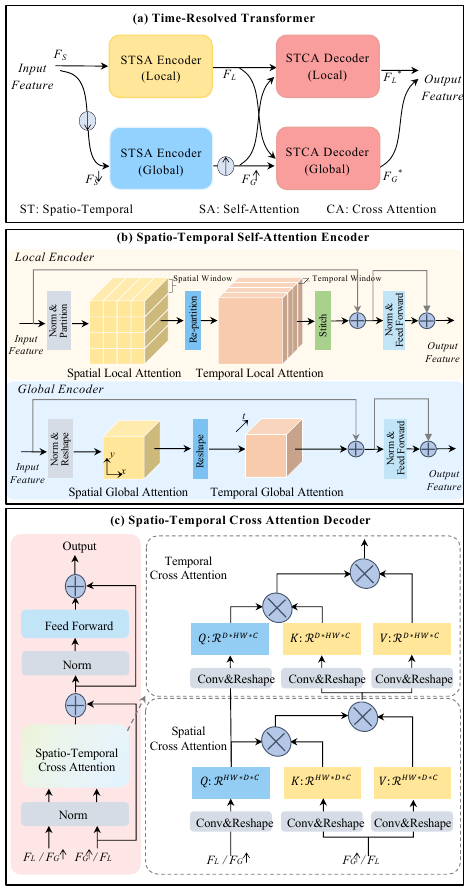}
	\captionsetup{font={small}, justification=raggedright}
	\caption{(a) An overview of the proposed time-resolved transformer. The symbols ``$\downarrow$" and ``$\uparrow$" with a circular block denote the downsampling and upsampling operators along the spatial dimension. The subscript $S$, $L$, and $G$ represent the shallow, local and global feature, while the superscript $*$ indicates the deep features. (b) An overview of spatio-temporal self-attention encoder. (c) An overview of spatio-temporal cross attention decoder.``$\times$" with a circle denotes the matrix multiplication.}
\label{fig:overview}
\end{figure}

To handle the challenges when applying transformers directly to 3D reconstruction from spatio-temporal transient measurements, we propose the time-resolved transformer (TRT).
The overview of TRT is illustrated in Fig.~\ref{fig:overview}(a). 
Given the shallow feature $F_S$, two spatio-temporal self-attention (STSA) encoders exploit the local and global correlations within
the spatio-temporal features, respectively, generating the local features $F_L$ and the global features $F_G$. After that, two spatio-temporal cross attention (STCA) decoders integrate the complementary local and global features, respectively, producing the deep local features $F_L^*$ and the deep global features $F_G^*$ with improved representation capabilities. Finally, the deep features $F_L^{*}$ and $F_G^{*}$ are fused to facilitate the subsequent 3D reconstruction of the target scene.

\noindent \textbf{Spatial-Temporal Self-Attention Encoder.}
To exploit the local correlation in transient measurement, we design a local spatio-temporal encoder based on the multi-head self-attention (MSA) mechanism. As shown in Fig.~\ref{fig:overview}(b), the local encoder consists of a spatial window-based MSA layer (W$_{s}$-MSA), a temporal window-based MSA layer (W$_{t}$-MSA), and a feed-forward network (FFN). Given the shallow features $F_S$ previously extracted, the local encoder first partitions the features into patches (with a size of $P_{s}^{2} \cdot T$, a number of $N_{s}=HW/P_{s}^{2}$) along spatial dimensions and processes these patches with W$_{s}$-MSA, individually. Then, the output features are reshaped and partitioned into patches (with a size of $P_{t} \cdot HW$, a number of $N_{t}= T/P_{t}$) along the temporal dimensions again and processed by W$_{t}$-MSA, individually. Finally, the output features are stitched and fed into the FFN, generating the features with local information. This process can be modeled as
\begin{equation}
	F_{L} = FFN\{W_{t}\mbox{-}MSA\{W_{s}\mbox{-}MSA\{{F}_{S}\}\}\},
\end{equation}
where $F_{L}$ $\in \mathbb{R}^{H\times W \times T \times C}$ denotes the output local features. By partitioning the input features into patches and extracting information within patches along spatial and temporal dimensions successively, the local encoder maintains the continuity of depth and intensity in a local region of the 3D transient measurement, which helps to provide more details for the reconstructions of hidden scenes.

To exploit the global correlations in transient measurements, we design a global spatio-temporal encoder based on the MSA mechanism. As shown in Fig.~\ref{fig:overview}(b), the global encoder consists of a full spatial MSA layer (F$_{s}$-MSA), a full temporal MSA layer (F$_{t}$-MSA), and an FFN. Given the shallow features $F_{S}$, the global encoder first downsamples the features along the spatial dimensions and processes the features with F$_{s}$-MSA. Then, the output features are reshaped and processed with F$_{t}$-MSA along the temporal dimension. Finally, the output features are fed into the FFN, generating the features with global information. This process can be modeled as
\begin{equation}
	F_{G} = FFN\{F_{t}\mbox{-}MSA\{F_{s}\mbox{-}MSA\{{F}_{S}^{\downarrow}\}\}\},
\end{equation}
where \textit{$F_{G}$} $\in \mathbb{R}^{\frac{H}{S}\times \frac{W}{S} \times T \times C}$ denotes the output global features. By downsampling the input features to a smaller scale and extracting information within the whole feature space along spatial and temporal dimensions successively, the global encoder maintains the consistency of depth and intensity in the whole 3D transient measurements, which helps to recover hidden scenes with large depth ranges and complicated geometries.

As demonstrated in the ablation study in Sec.~\ref{aba:sp_all} and Sec.~\ref{abla_spa}, our elaborate STSA encoders effectively capture the local and global correlations in transient measurements, which improves the reconstruction performance for challenging real-world scenes.

\noindent \textbf{Spatial-Temporal Cross Attention Decoder.}
To integrate both local and global information, we further design a spatio-temporal cross attention (STCA) decoder to improve the feature representation capability. 
The decoder consists of a local branch and a global branch based on our devised STCA mechanism, as shown in Fig.~\ref{fig:overview}(c). Both local and global branches contain an STCA layer, and an FFN interleaved with normalization. For the local branch, the local features $F_{L}$ and the upsampled global features $F_{G}^{\uparrow}$ (with the same scale as $F_{L}$) are fed into the STCA and FFN in sequence, generating the deep local features $F^{*}_{L}$ $\in \mathbb{R}^{H\times W \times T \times C}$ as
\begin{equation}
    \begin{split}
        F_{L}^{*} = &FFN\left \{STCA\left [Q, K, V\right ] \right \}, \\[-2pt]
	Q&= F_{G}^{\uparrow}, K=V=F_{L}.
    \end{split}
\end{equation}    	
For the global branch, the upsampled global features $F_{G}^{\uparrow}$ and the local features $F_{L}$ are fed into STCA and FFN in sequence generating the deep global features $F^{*}_{G}$ $\in \mathbb{R}^{H\times W \times T \times C}$ as
\begin{equation}
\begin{split}
	F_{G}^{*} = &FFN\left \{STCA\left [Q, K, V\right ] \right \},  \\[-2pt]
	Q&= F_{L}, K=V= F_{G}^{\uparrow}.
\end{split}
\end{equation}

As shown in Fig.~\ref{fig:overview}(c), STCA integrates the local and global features in a 3D token space, where local and global features are adopted as the query in turn. Given the two input features, a 1$\times$1$\times$1 convolution is conducted to produce the query (Q), key (K), and value(V), respectively. The space of Q, K, and V is reshaped to $\mathbb{R}^{HW \times D \times C}$ to calculate the spatial cross attention by matrix multiplication. After that, the output features are fed into a 1$\times$1$\times$1 convolution resulting a new K, and a new V. The space of the initial Q, the new K and the new V is reshaped to $\mathbb{R}^{D \times HW \times C}$ for calculating the temporal cross attention by matrix multiplication as well. As such, the two input features are integrated, and the local and global information complement each other simultaneously. The integration greatly improves the representation capability of the output features and promotes the reconstruction performance, as demonstrated in Sec.~\ref{aba:sp_all} and Sec.~\ref{abla_int}.


\section{Line-of-Sight Imaging}
In this section, we apply the proposed TRT for line-of-sight (LOS) single-photon imaging. We begin by outlining the forward model of the imaging process, which is also formulated for synthetic data simulation. Next, we introduce the proposed algorithm, TRT-LOS, along with the associated loss function tailored for LOS imaging. Finally, we present experimental details, results, and analysis, including a comprehensive evaluation of synthetic and real-world transient measurements from various imaging systems.
\subsection{Forward Model}
The ToF-based LOS imaging system primarily consists of a laser source, a time-resolved SPAD detector and a TCSPC. The system prototype is shown in Fig.~\ref{fig:correlations}(a). At time $t=0$, the laser emits short periodic pulses. After propagating to the object located at a distance $z$, a few photons are reflected back to the detector. Finally, TCSPC records the arriving photons over discrete time bins of duration $\Delta_t$. The detected photons, $\tau$, during the $n$-th time interval is
\begin{equation}
\label{eq:eq0}
\begin{split}
  \tau[n] &= \int_{n\Delta_t}^{(n+1)\Delta_t}(g*j)(t-\frac{2z}{c}) \mathrm{d}t,
\end{split}
\end{equation}
where $g$ and $j$ denote the shape of the light pulse and the corresponding jitter, and $c$ is the speed of light.

The photon detection is modeled as a Poisson process. After being captured by the detector with $N$ periods, the detected photons are represented by the temporal histogram
\begin{equation}
	\label{eq01}
	{H}[n] \sim \text { Poisson }(N\eta\phi\tau[n]+B),
\end{equation}
where $\eta$ is quantum efficiency, $\phi$ denotes attenuation factors including radial falloff and reflectance, $B$ represents the detected ambient light, and the scalar dark count. 
\begin{figure}[!t]
	\centering
	\includegraphics[width=0.48\textwidth]{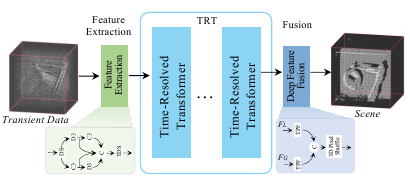}
	\caption{The flowchart of our proposed TRT-LOS. ``C" and ``D" with rectangular blocks denote the 3D convolution and 3D dilated convolution, respectively, with their kernel sizes behind. ``C" with circular blocks denotes the concatenation. ``DS" and ``TDS" denotes the downsampling operators along the spatio-temporal and temporal dimension, respectively. ``TPF" with a rectangular block denotes the pixel shuffle as the upsampling operator along the temporal dimension.} 
	\label{fig:sp_framework}
\end{figure}

\subsection{Network Architecture}
\noindent
\textbf{Overview.} Building on TRT, we propose an embodiment for LOS reconstruction, termed TRT-LOS, which effectively exploits both local and global correlations within the transient measurements. The proposed method, as illustrated in Fig.~\ref{fig:sp_framework}, consists of three main components: the feature extraction module, TRT blocks, and the deep-feature fusion module. Initially, the input transient measurement is processed by the feature extraction module. The extracted shallow features are then passed through the TRT blocks, where they are transformed into deep local and global features with enhanced representational power. Finally, the deep features are combined in the deep-feature fusion module to reconstruct the 3D volume of the target objects.

\noindent
\textbf{Feature Extraction.}
Given the input transient measurement, the high-dimensional measurement is first downsampled along the spatial and temporal axises. To expand the receptive field without increasing the parameters or computation cost, we employ a combination of interlaced 3D convolutions and 3D dilated convolutions. The extracted features are concatenated to capture rich contextual information from the spatio-temporal data. Subsequently, the enhanced features are processed by a temporal downsampling operator before being passed to the subsequent TRT blocks.

\noindent
\textbf{Deep Feature Fusion.}
After processing with the proposed TRT, the deep local features $F_L$ and deep global features $F_G$ are extracted, capturing rich spatio-temporal information. These features are then passed to the deep-feature fusion module. To mitigate interpolation artifacts, we first apply a temporal pixelshuffle operation to upsample the temporal dimension of the deep features. Subsequently, the upsampled deep local and global features are concatenated along the channel dimension. To further enhance spatio-temporal resolution, a 3D pixelshuffle operation is employed across both spatial and temporal dimensions. Different with interpolation, which may blur or lose critical spatial details, pixelshuffle preserves fine-grained details in the reconstructed 3D data, thereby improving both local and global feature representations in the upsampled output. The final predicted histogram is then modeled as
\begin{equation}
      H = 3DPF[CAT(TPF(F_L),TPF(F_G))],
\end{equation}
where $CAT$ means the concatenation, $3DPF$ and $TPF$ denote the 3D pixelshuffle and temporal pixelshuffle, respectively. The depth map of the corresponding target scene can be obtained by finding the maximum bin index along the temporal axis. However, we transform the max operation into a weighted probability summation for a differentiate output. The depth map of the target scene is generated from the predicted high-quality transient measurement by soft argmax operation~\cite{lindell2018single,peng2020photon}. The differentiate predicted depth map is formulated as
 \begin{equation}
    D = \sum_nn\cdot\text{softmax}(H[n]),
\end{equation}
where $n$ denotes the bin index of the histogram and the softmax function is operated along the temporal axis.

\noindent
\subsection{Loss Function}
The loss function contains two folds. The first is the Kullback-Leibler (KL) divergence between the predicted histogram $H^{(k)}$ and ground-truth histogram $\hat{H}^{(k)}$, which is formulated as follows
\begin{equation}
    \mathcal{L}_{KL}(H^{(k)},\hat{H}^{(k)}) =  \sum_{n}H^{(k)}[n]\cdot\text{log}\frac{H^{(k)}[n]}{\hat{H}^{(k)}[n]},
\end{equation}
where $k$ represents the spatial location and $n$ denotes the histogram index. To denoise the 2D depth map, we further introduce the total variation (TV) regularization item
\begin{equation}
\mathcal{L}_{TV} = \sum_k||D^{k+1}-D^{k}||_1.   
\end{equation}

The total loss function is thus given as
\begin{equation}
    \mathcal{L}=\mathcal{L}_{KL}(H^{(k)},\hat{H}^{(k)})+\gamma\mathcal{L}_{TV},
\end{equation}
where the hyperparameter $\gamma$ weights the TV loss.
\subsection{Experiments on Simulated Data}

\setlength{\tabcolsep}{10pt}
\begin{table*}[!t]
\small
\begin{center}
\caption{Quantitative comparisons of different methods on the test set which is simulated from Middlebury2014 dataset.}
\label{table:sp_qua}
\begin{tabular}{ccccccccc}
\hline
SBR & LM~\cite{bar1969communication} & Shin~\cite{shin2016photon} & Rapp~\cite{rapp2017few} & CAPSI~\cite{lee2023caspi} & Lindell~\cite{lindell2018single} & Lindell\_I~\cite{lindell2018single} & Peng~\cite{peng2020photon} & TRT-LOS \\ \hline
10:2 & 0.8230 & 0.1167 & 0.0985 & 0.0762 & 0.0716 & 0.0452 & 0.0367 &  \textbf{0.0289} \\
5:2 & 1.7218 & 0.1321 & 0.1004 & 0.1106 & 0.0876 & 0.0567 & 0.0429 &  \textbf{0.0333} \\
2:2 & 3.3486 & 0.5066 & 0.1265 & 0.1776 & 0.1213 & 0.0753 & 0.0560 &  \textbf{0.0461} \\ \hline
Avg. & 2.0103 & 0.2518 & 0.1084 & 0.1215 & 0.0935 & 0.0591 & 0.0442 & \textbf{0.0361} \\ \hline
10:10 & 1.1921 & 0.2516 & 0.0910 & 0.0748 & 0.0671 & 0.0462 & 0.0378 &  \textbf{0.0318} \\
5:10 & 2.3576 & 1.3158 & 0.1011 & 0.1308 & 0.0807 & 0.0569 & 0.0459 &  \textbf{0.0386} \\
2:10 & 4.2354 & 3.2273 & 0.1581 & 0.1992 & 0.1125 & 0.0774 & 0.0637 &  \textbf{0.0547} \\
\hline
Avg. & 2.5951 & 1.5982 & 0.1167 & 0.1349 & 0.0868 & 0.0602 & 0.0492 & \textbf{0.0417} \\ 
\hline
10:50 & 1.6416 & 3.2351 & 0.0947 & 0.1000 & 0.0612 & 0.0546 & 0.0439 &  \textbf{0.0371} \\
5:50 & 3.1882 & 3.9722 & 0.1215 & 0.1878 & 0.0718 & 0.0690 & 0.0554 &  \textbf{0.0475} \\
2:50 & 5.1155 & 4.4203 & 0.2508 & 0.4025 & 0.1161 & 0.1036 & 0.0819 &  \textbf{0.0679} \\
\hline
Avg. & 3.3151 & 3.8759 & 0.1557 & 0.2301 & 0.0830 & 0.0757 & 0.0604 & \textbf{0.0508} \\
\hline
3:100 & 4.8436 & 4.4960 & 0.2314 & 0.4319 & 0.4797 & 0.1007 & 0.0779 &  \textbf{0.0652} \\
2:100 & 5.4849 & 4.5711 & 0.3680 & 0.7923 & 0.6978 & 0.1227 & 0.0930 &  \textbf{0.0749} \\
1:100 & 6.0731 & 4.6467 & 0.7793 & 2.3330 & 1.4402 & 0.2641 & 0.1380 &  \textbf{0.1038} \\
\hline
Avg. & 5.4672 & 4.5713 & 0.4596 & 1.1857 & 0.8726 & 0.1625 & 0.1030 & \textbf{0.0813} \\ 
\hline
\end{tabular}
\end{center}
\end{table*}
\subsubsection{Data Simulation and Evaluation Metric}
\begin{figure}[!t]
	\centering
	\includegraphics[width=0.48\textwidth]{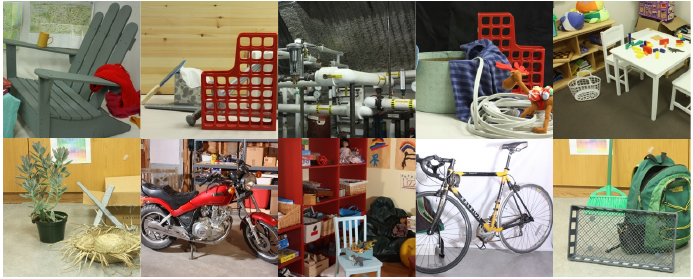}
	\caption{Thumbnails of the synthetic test scenes from Middlebury2014 dataset.} 
	\label{fig:middbury}
\end{figure}

The spatial resolution of previous synthetic datasets was set to 64$\times$64~\cite{lindell2018single,peng2020photon}, which is relatively small due to the computational limitations at the time. The coarse-grained spatial features in these datasets make it challenging for the model generalize to complex scenes. Thus, we introduce a new synthetic dataset with 256$\times$256 spatial resolution. For the large-scale training dataset, we utilize the RGB-D dataset NYU v2~\cite{silberman2012indoor} for simulation, according to the forward model Eq.~\ref{eq01}. The measurement has a resolution of 256$\times$256$\times$1024 with a bin width of 80 ps. For the signal and background noise level, the average detected signal photons per pixel is randomly set to 2, 5, and 10, and the corresponding background noise is randomly set to 2, 10, and 50. Besides, we also introduce the extreme SBRs in the training dataset, i.e., 1:100, 2:100, and 3:100. A total of 13051 and 2742 samples are generated for training and validation. For the test set, we select Middlebury2014~\cite{scharstein2014high}, which features more complex scenarios and provides accurate ground-truth depth maps. A total of 10 scenes, each containing 12 SBRs, are generated. The scenes are presented in Fig.~\ref{fig:middbury}. The evaluation metric is root mean square error (RMSE) for the predicted depth maps.
\setlength{\tabcolsep}{12pt}
\begin{figure*}[!t]
	\centering
	\includegraphics[width=\textwidth]{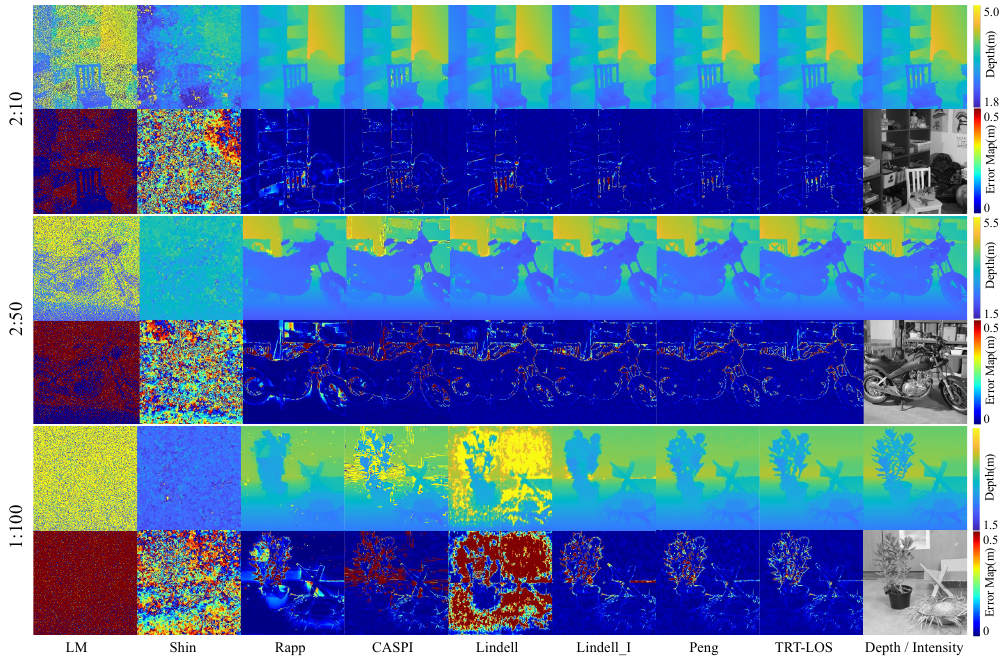}
	\caption{Reconstructed results from the simulated test set under different SBR conditions. The odd and even rows are the depth maps and depth error maps, respectively. The last column lists the ground-truth depth map and intensity image. The color bars show the value of depth and the error map..}
\label{fig:spsyn1} 
\end{figure*}
\subsubsection{Implementation Details}
We implement our method using PyTorch~\cite{paszke2019pytorch} and train the models on the simulated data for 50 epochs with a batch size of 1 and a learning rate of $10^{-4}$. For optimization, we use the AdamW~\cite{loshchilov2018decoupled} solver. The hyper-parameter $\gamma$ is set to $10^{-5}$. 
All the experiments are conducted on a computer with 4 NVIDIA GeForce 3090 GPUs. The comparisons are conducted with the state-of-the-art baselines, including computational methods: LM~\cite{bar1969communication}, Shin~\cite{shin2016photon}, Rapp~\cite{rapp2017few}, CAPSI~\cite{lee2023caspi},
and deep-learning-based methods: Lindell~\cite{lindell2018single} (trained without intensity images), Lindell\_I~\cite{lindell2018single} (trained with intensity images)  and Peng~\cite{peng2020photon}. The implementations of the baseline methods follow their publicly available codes. 
For inference, we perform the test using the full resolution rather than processing it patch by patch. All baselines follow this stratgy, except for~\cite{lindell2018single}, which requires a large amount of GPU memory when using high-resolution inputs. As a result, inference is performed on one-quarter of the input at a time, and the results are then stitched together for~\cite{lindell2018single}.
\subsubsection{Simulated Results}
The quantitative results in terms of the reconstructed depth maps on the test set are listed in Table~\ref{table:sp_qua}. We list the SBRs at various signal levels and provide the average values. It can be observed that our method achieves the best performance under all conditions. For the computational methods, they perform well under high SBR conditions, but their performance experiences a sharp decline as noise increases, while Rapp exhibits good robustness. Starting from the deep learning algorithm Lindell\_I, the performance shows a significant leap, far surpassing traditional methods in depth estimation metrics. Although the Lindell method without intensity data shows a decline in performance compared to Lindell\_I, most metrics still outperform traditional methods. With the introduction of a finely designed network for single-photon data, the performance of Peng improves notably compared to Lindell\_I, indicating that modeling non-local information of single-photon data benefits 3D reconstruction. As we further explore the local and global correlations within the transient measurements, our method achieves the best across all levels, demonstrating the effectiveness of TRT-LOS.

As shown in Fig.~\ref{fig:spsyn1}, we provide qualitative visualizations, including depth maps and corresponding error maps. It can be observed that the computational algorithms LM and Shin perform poorly, only managing to recover rough outlines with significant overall errors. The other two computational algorithms, Rapp and CASPI, perform well. CASPI excels in detail recovery compared to Rapp but struggles with background recovery, which leads to its quantitative metrics being lower than Rapp’s. The deep learning methods achieve good reconstruction results for both the foreground and background. Compared to Lindell, Peng shows smaller errors on a global scale. Our method further demonstrates robustness in detailed recovery, such as the backrest bars of a chair and the brake lines of a motorcycle, highlighting the superior performance of TRT-LOS. In situations with sparse effective signals and extremely high noise level as in the last row of Fig.~\ref{fig:spsyn1}, our method, however, still maintains a clear advantage in recovering difficult details, such as branches.


\setlength{\tabcolsep}{4.5pt}
\begin{table}[!t]
\small
\begin{center}
\caption{Ablation studies on loss items and deep feature fusion. ``Loc'' and ``Glo'' represent the deep local and global features. ``KL'' and ``TV'' denote the Kullback-Leibler and total variation loss items. The
quantitative results are average for each noise group, i.e., X=2, 5, 10 and Y=1, 2, 3.}
\label{table:asd}
\begin{tabular}{cc|cc|cccc}
\hline
Loc & Glo & KL & TV & X:2   & X:10    & X:50    & Y:100   \\ \hline

\checkmark   & \checkmark   & \checkmark  & ×   &        0.0370 & 0.0439& 0.0546& 0.0876\\
\checkmark   & \checkmark   & \checkmark  & \checkmark    & \textbf{0.0361} & \textbf{0.0417} & \textbf{0.0508} & \textbf{0.0813} \\
\hline
×   & \checkmark   & \checkmark  & \checkmark    & 4.7183 &  4.6533& 4.7042& 4.6542\\
\checkmark   & ×   & \checkmark  & \checkmark    &       0.0389& 0.0452& 0.0563& 0.0887\\ 
\checkmark   & \checkmark   & \checkmark  & \checkmark    & \textbf{0.0361} & \textbf{0.0417} & \textbf{0.0508} & \textbf{0.0813} \\
\hline
\end{tabular}
\end{center}
\vspace{-0.5cm}
\end{table}

\setlength{\tabcolsep}{5pt}
\begin{table}[!t]
\small
\begin{center}
\caption{Ablation studies on spatio-temporal attention and the integration of local-global features. ``T.'' and ``S.'' denote that the attention mechanism only operates on the temporal or  spatial dimension, respectively. 
The quantitative results are average for each noise group, i.e., X=2, 5, 10 and Y=1, 2, 3.}
\label{table:sp_st_int}
\begin{tabular}{cc|c|cccc}
\hline
T. & S. & Integration & X:2     & X:10    & X:50    & Y:100   \\ \hline
×    & \checkmark & LGInt(Ours)       & 0.0994 & 0.1956 & 0.3148 & 0.4200\\
\checkmark    & ×   & LGInt(Ours)      & 0.0392 & 0.0445& 0.0556 & 0.0930 \\ 
\checkmark    & \checkmark   & LGInt(Ours)      & \textbf{0.0361} & \textbf{0.0417} & \textbf{0.0508} & \textbf{0.0813} \\
\hline
\checkmark    & \checkmark   & NoInt      & 0.0367& 0.0439 & 0.0554& 0.0892 \\
\checkmark    & \checkmark   & LocInt      & 0.0361 & 0.0420& 0.0511& 0.0816\\
\checkmark    & \checkmark   & GloInt      & 0.0361 & 0.0420& 0.0511& 0.0816\\
\checkmark    & \checkmark   & LGInt(Ours)      & \textbf{0.0361} & \textbf{0.0417} & \textbf{0.0508} & \textbf{0.0813} \\ \hline
\end{tabular}
\end{center}
\end{table}

\subsubsection{Ablation Studies}
\label{aba:sp_all}
We conduct ablation studies to assess the effectiveness of the loss component, deep feature fusion, spatio-temporal attention, and the integration of local-global features. Additionally, we investigate the impact of the number of the TRT blocks and the input channels of TRT.

\noindent
\textbf{Loss Item.} 
As discussed above, we introduce the total variation regularization item for denoising the predicted depth map. To explore the effectiveness, we disable this loss item. As can be seen in Table~\ref{table:asd}, the quantitative depth metrics increase.  

\noindent
\textbf{Deep Feature Fusion.} 
After TRT blocks, the deep local and global features are integrated for the features with high capabilities. The experimental results are presented in Table~\ref{table:asd}. When the method only contains global features, the method perform quite worse, which is caused by low resolution from downsampling operation in the global branch. When the local features are introduced, the depth metrics decrease significantly, highlighting the contribution of the local features in single-photon measurements. Finally, when all features are fused, the method achieve best on all scenarios with various SBRs, demonstrating the effectiveness of the local and global feature fusion.

\setlength{\tabcolsep}{10pt}
\begin{table}[!t]
\small
\begin{center}
\caption{Ablation studies on the number of TRT blocks.}
\label{table:abla_num}
\begin{tabular}{ccccc}
\hline
SBR & 1 & 6 & 12(Ours) & 24 \\
\hline
10:2 & 0.0342 & 0.0285 & 0.0289 & 0.0282 \\
5:2 & 0.0403 & 0.0329 & 0.0333 & 0.0323 \\
2:2 & 0.0544 & 0.0472 & 0.0461 & 0.0468 \\
\hline
Avg. & 0.0430 & 0.0362 & 0.0361 & \textbf{0.0360} \\
\hline

10:10 & 0.0383 & 0.0316 & 0.0318 & 0.0314 \\
5:10 & 0.0460 & 0.0391 & 0.0386 & 0.0390 \\
2:10 & 0.0641 & 0.0568 & 0.0547 & 0.0582 \\
\hline
Avg. & 0.0495 & 0.0425 & \textbf{0.0417} & 0.0429 \\
\hline

10:50 & 0.0484 & 0.0388 & 0.0371 & 0.0386 \\
5:50 & 0.0592 & 0.0495 & 0.0475 & 0.0507 \\
2:50 & 0.0791 & 0.0698 & 0.0679 & 0.0726 \\
\hline
Avg. & 0.0623 & 0.0527 & \textbf{0.0508} & 0.0540 \\
\hline

3:100 & 0.0776 & 0.0681 & 0.0652 & 0.0708 \\
2:100 & 0.0866 & 0.0778 & 0.0749 & 0.0799 \\
1:100 & 0.1126 & 0.1048 & 0.1038 & 0.1073 \\
\hline
Avg. & 0.0922 & 0.0836 & \textbf{0.0813} & 0.0860 \\
\hline

\end{tabular}
\end{center}
\end{table}

\begin{figure*}[!t]
	\centering
	\includegraphics[width=\textwidth]{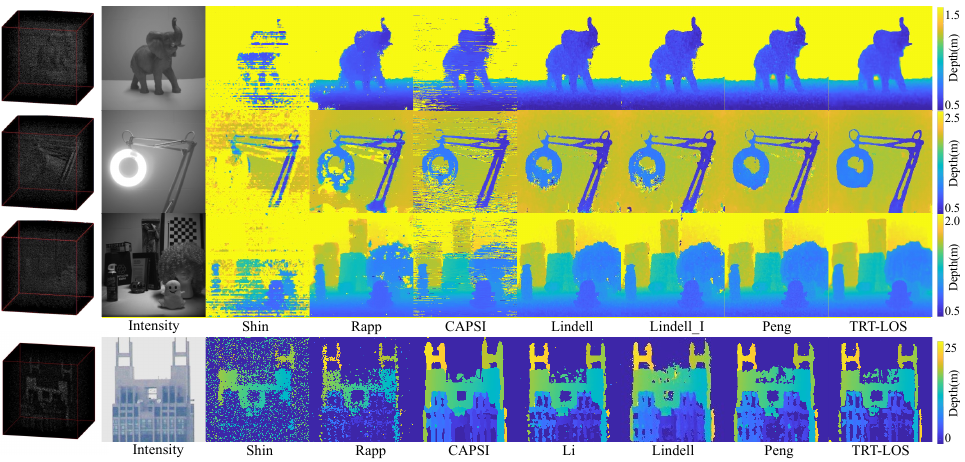}
 	\captionsetup{font={small}, justification=raggedright}
	\caption{Reconstruction results of the methods for real-world scenes. The first three rows are captured by the indoor line-of-sight imaging system~\cite{lindell2018single} and the last scenario is captured by the outdoor line-of-sight imaging prototype~\cite{li2020single}.}
\label{fig:sp_rw_lindell} 
\end{figure*}

\setlength{\tabcolsep}{10pt}
\begin{table}[!t]
\small
\begin{center}
\caption{Ablation studies on the input channels of TRT.}
\label{table:abla_chan}
\begin{tabular}{ccccc}
\hline
SBR & 16 & 32 & 64(Ours) & 128 \\
\hline
10:2 & 0.0365 & 0.0307 & 0.0289 & 0.0304 \\
5:2 & 0.0410 & 0.0352 & 0.0333 & 0.0347 \\
2:2 & 0.0527 & 0.0479 & 0.0461 & 0.0470 \\
\hline
Avg. & 0.0434 & 0.0380 & \textbf{0.0361} & 0.0374 \\
\hline
10:10 & 0.0396 & 0.0335 & 0.0318 & 0.0326 \\
5:10 & 0.0455 & 0.0403 & 0.0386 & 0.0396 \\
2:10 & 0.0608 & 0.0575 & 0.0547 & 0.0553 \\
\hline
Avg. & 0.0486 & 0.0437 & \textbf{0.0417} & 0.0425 \\
\hline
10:50 & 0.0457 & 0.0397 & 0.0371 & 0.0383 \\
5:50 & 0.0550 & 0.0500 & 0.0475 & 0.0479 \\
2:50 & 0.0742 & 0.0703 & 0.0679 & 0.0681 \\
\hline
Avg. & 0.0583 & 0.0533 & \textbf{0.0508} & 0.0514 \\
\hline
3:100 & 0.0747 & 0.0697 & 0.0652 & 0.0647 \\
2:100 & 0.0848 & 0.0798 & 0.0749 & 0.0757 \\
1:100 & 0.1462 & 0.1084 & 0.1038 & 0.1064 \\
\hline
Avg. & 0.1019 & 0.0860 & \textbf{0.0813} & 0.0822\\
\hline
\end{tabular}
\end{center}
\end{table}

\noindent
\textbf{Spatio-Temporal Attention.} 
\label{abla_st_int}We design STSA and STCA mechanisms to capture both local and global correlations in 3D transient measurements across spatial and temporal dimensions. To evaluate the efficiency of spatial and temporal attention individually, we conduct experiments with the self-attention encoder and cross attention decoder, with results presented in Table ~\ref{table:sp_st_int}. As shown, spatial and temporal attentions have distinct contributions to reconstruction performance, and combining both leads to an improvement in overall performance.

\noindent
\textbf{Local-Global Feature Integration.} 
In the cross attention decoder, local and global features are integrated by allowing them to query each other in the token space. To assess the effectiveness of this integration (LGInt), we compare it with alternative approaches: NoInt (no integration between the two branches), LocInt (integration only within the local branch), and GloInt (integration only within the global branch). When no integration occurs between the local and global features, the method performs the worst. However, when integration occurs within either the local or global branch alone, performance improves. The best performance is achieved when all features are fully integrated as shown in Table ~\ref{table:sp_st_int}.

\noindent
\textbf{Numbers and Channels of TRT.} 
To further explore the impact of both the number of TRT blocks in the proposed method and the input channels within each TRT, we conduct separate ablation studies. The quantitative results are presented in Table~\ref{table:abla_num} and Table~\ref{table:abla_chan}. Regarding the number of TRT blocks, performance improves as the number increases; however, this trend does not hold when the number of TRT blocks reaches 24. Specifically, the method with 24 TRT blocks performs best in scenes with high SBRs, but its performance deteriorates as the noise level increases. This phenomenon is consistent with the number of channels in the TRT. From the quantitative results presented in Table~\ref{table:abla_chan}, we observe that when the number of input channels is 64, the network achieves the best performance under all SBR conditions. 
\subsection{Experiments on Real-world Data}
\noindent
\textbf{Data Preparation.}
To evaluate the generalization capability of the proposed method, we conduct tests on real-world transient measurements captured by different imaging systems. First, we use the real-world dataset from~\cite{lindell2018single}, where the transient measurements, with a spatial resolution of 256$\times$256, are captured in an indoor environment. The distances between the prototype and target objects range from 0.5\textit{m} to 2.5\textit{m}. Next, we test the models on measurements obtained from the long-distance outdoor imaging system~\cite{li2020single}, where the scene $\textit{K11}$ building is located 21\textit{km} away, with a spatial resolution of 128$\times$128. During inference, the input transient measurements are processed at their full spatial resolution. Except for Lindell and Lindell\_I, which encountered memory limitations, the measurements are cropped, and the reconstructed results are subsequently stitched together. Due to the infinite depth of the background in~\cite{li2020single}, the final predicted depth map is filtered using a constant threshold which is calculated as the mean of the predicted reflectivity image.

\noindent
\textbf{Real-world Results.} As illustrated in Fig.~\ref{fig:sp_rw_lindell}, we present the evaluation results of all models on real-world data. The traditional algorithms, LM and Shin, demonstrate suboptimal performance, failing to recover the structure and introducing significant noise. In contrast, the traditional methods, Rapp and CASPI, achieve better structural recovery, although they still produce substantial background noise. 
In the extra-long-range scenario captured by~\cite{li2020single}, we also evaluate the method referred to as Li. Although it yields clearer overall structures, its reconstruction exhibits deficiencies in boundary completeness.
When it comes to deep learning approaches, it is noteworthy that the models trained on our newly developed simulation dataset consistently outperform the results reported in the original paper when evaluated on the real-world dataset. Among all methods, our proposed TRT-LOS achieves the best performance. It not only reconstructs the overall scene with high fidelity and completeness, but also excels in recovering fine-grained details along object edges. Compared to other models, TRT-LOS demonstrates superior robustness in handling complex real-world noise, offering both global consistency and local precision.

\section{Non-Line-of-Sight Imaging}
In this section, we apply the proposed TRT to non-line-of-sight (NLOS) imaging. We begin by building the forward model of the imaging system, which is then utilized for synthetic data simulation. Next, we present the proposed algorithm, TRT-NLOS, and the corresponding loss function tailored for NLOS imaging. Finally, we provide a comprehensive overview of the experimental setup, along with the results and analysis, including both synthetic and real-world transient measurements captured by various imaging systems.
\subsection{Forward Model}
The ToF-based NLOS imaging system mainly contains a laser source, a time-resolved SPAD detector and a relay wall, shown in Fig.~\ref{fig:correlations}(b). 
The system works in a confocal manner, where the laser projects short periodic light pulses $\delta(t)$ toward the relay wall at illumination point $p_{l}$, from where the light is diffusely scattered at time $t=0$ and targets the hidden object. After integrated with the object at a certain target point $p_{t}$, a fraction of the light is reflected back to the relay wall after time interval $t$ and finally captured by the SPAD at the sampling point  $p_{s}$, resulting in a 3D spatio-temporal volume $\tau(p_{s},p_{t},t)$, known as transient measurement. The transient measurement, containing both geometric and photometric information of the hidden object, is a function of illuminated point $p_l$, sampling point $p_s$, and target point $p_t$, modeled as
\begin{equation}
\label{eq:eq1}
\begin{split}
  \tau(p_{s},p_{t},t) &= \iiint_{\Omega}\rho(p_{t})\cdot f(n_{l\to t},n_{t\to s})\cdot \varphi \\
  &\cdot \delta(r_{l}+r-tc) \mathrm{d}\Omega,
\end{split}
\end{equation}
where $\Omega$ denotes the spherical surface of scattered pulse light from the relay wall. $\rho(\cdot)$ denotes the
albedo of the hidden object. $n_{a\to b}$ means the normalized direction from point $a$ to point $b$. $f(\cdot)$ represents the bidirectional reflectance distribution function, containing diffuse, specular, and retroreflective components. $c$ is the speed of light. $r_l$ is the distance between the illumination point and the target point, while $r_s$ is the distance between the sampling point and the target point. $\varphi$ is the geometry radiometric term modeled as
\begin{equation}
\label{eq:eq2}
\varphi = \frac{(n_{s\to t} \cdot n_{t})\cdot \upsilon_{s\to t}}{r_{l} \cdot r},
\end{equation}
where $\upsilon_{s\to t}$ represents the visibility of the target point to the sampling point. 

This forward model is general and only assumes no inter-reflections in the hidden scene. Similar to ~\cite{o2018confocal,chopite2020deep,chen2020learned}, we model the photon detection of SPAD with an inhomogeneous Poisson process~\cite{snyder1975random} and store the transient measurement in the form of a histogram matrix ${H}[n]$ with discrete temporal bins. After being captured by the detector within N pulses, the discrete transient histogram ${H}[n]$ can be accumulated as
\begin{equation}
	\label{eq3}
	{H}[n] \sim \text { Poisson }(N\cdot\eta\tau(p_{s},p_{t},n^J) + B),
\end{equation}
where $n$ is the index of the temporal bins including modeling the jitter, and $B$ is the noise photon detections, including both background photons and dark counts~\cite{bronzi2015spad} of SPAD sensors.

\begin{figure*}[!t]
	\centering
	\includegraphics[width=1\textwidth]{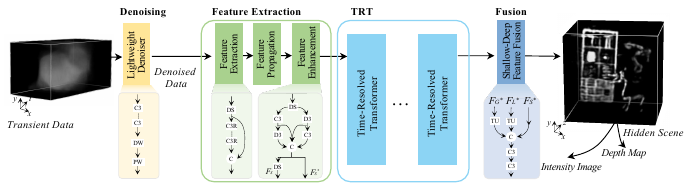}
	\caption{The flowchart of our proposed TRT-NLOS. ``DW" and ``PW" denote the depthwise convolution and the pointwise convolution. ``C" and ``D" with rectangular blocks denote the 3D convolution and 3D dilated convolution, respectively, with their kernel sizes behind. ``C" with circular blocks denotes the concatenation. ``DS" with a rectangular block denotes the downsampling operators along the temporal dimension. ``TU" with a rectangular block denotes the upsampling operator along the temporal dimension. ``$\downarrow$" and ``$\uparrow$" with a circular block denote the downsampling and upsampling operators along the spatial dimension.} 
	\label{fig:nlos_framework}
\end{figure*}



\subsection{Network Architecture}
\noindent
\textbf{Overview.}
Building on TRT, we propose an embodiment for NLOS reconstruction, termed as TRT-NLOS.
The framework is shown in Fig.~\ref{fig:nlos_framework}. Firstly, the input transient data is fed to a 
lightweight denoiser, which benefits the subsequent reconstruction. Then, the feature extractor extracts the shallow features $F_S$ and $F_S^{*}$ with physics-based priors. After TRT, the deep local features $F_L^*$ and the deep global features $F_G^*$ are generated with improved representation capabilities. Finally, the shallow features $F_S^{*}$ and the deep features $F_L^{*}$ and $F_G^{*}$, are fused together to reconstruct the 3D volume of hidden objects, generating the intensity image and the depth map.

\noindent
\textbf{Transient Measurement Denoiser.}
The quality of reconstruction results is determined by both the performance of the reconstruction network and the quality of the input data. To mitigate the impact of noisy data, we first introduce a lightweight denoising head prior to the reconstruction process, ensuring clean, high-quality transient measurements. This step significantly enhances the performance of the subsequent reconstruction. The initial part of the denoiser consists of four standard 3D convolutions, each followed by a ReLU activation. As the number of channels increases, the computational cost and parameter count rise substantially due to the 3D spatio-temporal input. To address this, the latter part of the denoiser employs depthwise separable convolutions, which reduce the computational burden while maintaining high performance.

\noindent
\textbf{Shallow Feature Extraction.}
The shallow feature extractor consists of a feature extraction layer, a feature transform layer, and a feature enhancement layer, as shown in Fig.~\ref{fig:nlos_framework}. Given the denoised data, we first downsample the input in the temporal dimension and extend the channel dimension by several residual convolutions with the feature extraction layer. Inspired by the existing learning-based methods~\cite{chen2020learned,MufuICCP}, we transform the spatio-temporal features to the 3D spatial domain with a physics-based prior FK~\cite{lindell2019wave} in the feature transform layer. We further enhance the output features by several interlaced 3D convolutions and 3D dilated convolutions to enlarge the receptive field with the feature enhancement layer, producing the shallow features \textit{$F_{S}^{*}$} and \textit{$F_{S}$}, where \textit{$F_{S}$}$\in \mathbb{R}^{H\times W \times T \times C}$. $H$, $W$, $T$, and $C$ denote the height, width, time, and channel dimension of the feature volume.

\noindent
\textbf{Shallow-Deep Feature Fusion.}
Finally, the deep local features $F_L^{*}$ and deep global features $F_G^{*}$ are fed to 3D deconvolutions to upsample the temporal dimension (with the same scale as $F_{S}^{*}$). Then the upsampled deep local and global features, and the shallow features $F_{S}^{*}$ are concatenated along the channel dimension and then fused with 3D convolutional layers to reconstruct the 3D volume $V$ of the hidden scene. The intensity image $\hat{I}$ is obtained with a max operator along the $z$ axis, while depth map $\hat{D}$ is obtained with an argmax operator along the $z$ axis, which can be modeled as 
\begin{equation}
    \begin{split}
      V = FUS[CAT(F_S^*,F_L^{*\uparrow},F_G^{*\uparrow})],\\
    \hat{I} = \text{max}_z(V), \hat{D} = \text{argmax}_z(V),
    \end{split}
\end{equation}
where $CAT$ denotes the concatenation along the channel dimension, while $FUS$ contains several 3D convolutions layers. 

\subsection{Loss Function}
The loss function is threefold: the measurement loss $\mathcal{L}_{M}$, the intensity loss $\mathcal{L}_{I}$ and the depth loss $\mathcal{L}_{D}$. The first is defined as the Manhattan distance between the denoised measurement $\hat{\rho}$ and the ground-truth $\rho$. The second is defined as the Manhattan distance between the reconstructed intensity image $\hat{I}$ and the ground-truth $I$. The last is the Manhattan distance between the reconstructed depth map $\hat{D}$ and the ground-truth $D$. They are denoted as
\begin{equation}
\begin{split}
    \mathcal{L}_{M}(\rho,\hat{\rho}) =\| \rho - \hat{\rho} \|_{1}, \\
    \mathcal{L}_{I}(I,\hat{I}) =\| I - \hat{I} \|_{1},\\
    \mathcal{L}_{D}(D,\hat{D}) =\| D - \hat{D} \|_{1}.
\end{split}
\end{equation}

The final loss function to train the network is 
\begin{equation}
    \mathcal{L} = \mathcal{L}_{M}(\rho,\hat{\rho}) + \alpha \mathcal{L}_{I}(I,\hat{I}) + \beta \mathcal{L}_{D}(D,\hat{D}),
\end{equation}
where $\alpha$ and $\beta$ are weighting factors.


\subsection{Experiments on Simulated Data}
\subsubsection{Data Simulation and Evaluation Metrics}
\noindent
Following~\cite{chopite2020deep,chen2020learned,MufuICCP}, we simulate the training and testing data using the transient rasterizer~\cite{chen2020learned} with default settings. A total of 2903 transient measurements with corresponding gray-scale images are rendered from the motorcycles in the ShapeNet Core dataset~\cite{ChangFGHHLSSSSX15}. Each measurement has a spatial resolution of 256$\times$256$\times$512 with a bin width of 33 ps. A total of 2526 samples are adopted for training while the remaining 297 samples are used for testing, denoted as \textbf{Seen} testing data. 
To validate the generalization capability, we also render 526 transient measurements from other objects (i.e., baskets, helmets, cars, and so on), denoted as \textbf{Unseen} testing data.

The quantitative evaluation metrics are twofold. For the intensity image, we compute the peak signal-to-noise ratio (PSNR), and structural similarity metrics (SSIM) averaged on the corresponding test samples. For the depth map, we compute the root mean square error (RMSE) and mean absolute distance (MAD) averaged on the test samples. Considering the large amount of blank space in the background, we crop the test data based on the GT during evaluation, focusing only on the central and valid region. The approach ensures more robust and reasonable results.

\begin{figure*}[!t] 
\centering
\includegraphics[width=0.98\textwidth]{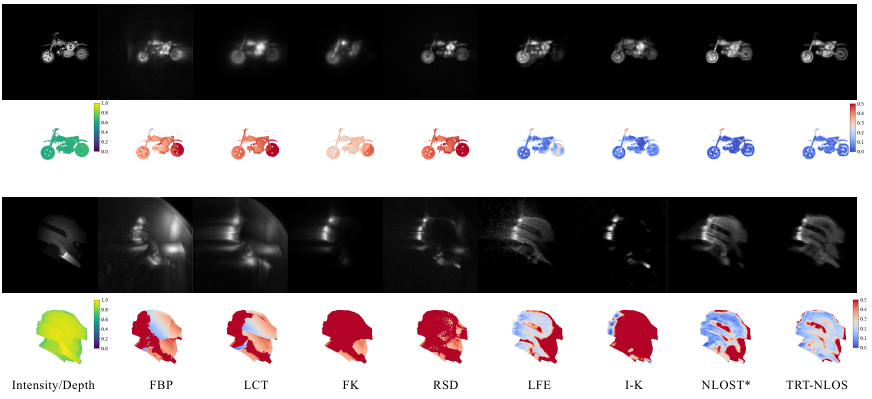}
\captionsetup{font={small}, justification=raggedright}
\caption{Reconstructed results from \textbf{Seen} (First scene) and \textbf{Unseen} (Second scene) test sets on the simulated datasets. The first row is the ground-truth intensity and depth. The odd and even rows are the intensity images and the depth error maps, respectively. The color bars show the value of depth and the error map.}
\label{fig:syn_all_int}
\end{figure*}

\setlength{\tabcolsep}{4.5pt}
\begin{table}[t]
\small
\begin{center}
\caption{Quantitative comparisons of different methods in terms of reconstructing intensity images and depth maps on the \textbf{Seen} and \textbf{Unseen} test set. The best in bold, the second in underline.}
\label{table:syn_all}
\begin{tabular}{c|c|cc|cc}
\hline  
  \multirow{2}{*}{Data} &\multirow{2}{*}{Methods} & \multicolumn{2}{c}{ Intensity} & \multicolumn{2}{c}{ Depth}\\
 \cline{3-6}
 & & PSNR$\uparrow$ & SSIM$\uparrow$ & RMSE$\downarrow$ & MAD$\downarrow$  \\
\hline
\multirow{10}{*}{\rotatebox{90}{Seen}}
&FBP~\cite{velten2012recovering}   & 19.96 & 0.1846 & 0.7053 &0.6694 \\
&LCT~\cite{o2018confocal}   & 19.51	 & 0.3615 & 0.4886 & 0.4639   \\
&RSD~\cite{liu2019non}   & 21.74	& 0.1817 & 0.5677 & 0.5320  \\
&FK~\cite{lindell2019wave}  & 21.69	& 0.6283  & 0.6072 & 0.5801  \\
&LFE~\cite{chen2020learned}   & 23.27&0.8118&0.1037&0.0488 \\
&I-K~\cite{yu2023enhancing}   & 23.44&0.8514&0.1041&0.0476 \\
&NLOST~\cite{li2023nlost}   & 23.74&0.8398&0.0902&0.0342 \\
&NLOST*~\cite{li2023nlost}   & \underline{24.03}&\underline{0.8583}&\underline{0.0849}&\textbf{0.0292} \\
&TRT-NLOS   & \textbf{24.15}&\textbf{0.8610}&\textbf{0.0836}&\underline{0.0319} \\
\hline  
\multirow{10}{*}{\rotatebox{90}{Unseen}}
&FBP~\cite{velten2012recovering}   & 17.86 & 0.1269 & 0.4860& 0.4453\\
&LCT~\cite{o2018confocal}   & 17.47	 & 0.1773 & 0.5562 & 0.5164   \\
&RSD~\cite{liu2019non}   & 19.31	& 0.1660 & 0.5162 & 0.4814  \\
&FK~\cite{lindell2019wave}  & 19.79	& 0.5242  & 0.7776 & 0.7751  \\
&LFE~\cite{chen2020learned}   & 21.28&0.6301&0.2883&0.1694 \\
&I-K~\cite{yu2023enhancing}   & 20.71&0.7513&0.2887&0.1694 \\
&NLOST~\cite{li2023nlost}   & 21.22&0.7254&\underline{0.2674}&\underline{0.1364} \\
&NLOST*~\cite{li2023nlost}  & \underline{21.30} &\underline{0.7643}&0.2766&0.1372 \\

&TRT-NLOS   & \textbf{22.54}&\textbf{0.8016}&\textbf{0.2486}&\textbf{0.1225} \\
\hline  
\end{tabular}
\end{center}
\end{table}

\subsubsection{Implementation Details}
We implement our method using PyTorch~\cite{paszke2019pytorch} and train the network on the simulated data with a batch size of 4. We initialize the network randomly and adopt the AdamW~\cite{loshchilov2018decoupled} solver with a learning rate of $10^{-4}$ and an exponential decay of 0.95. The hyper-parameter $\alpha$ and $\beta$ are set to $1$. 
We make comparisons with the existing baselines, including physics-based methods: FBP~\cite{velten2012recovering}, LCT~\cite{o2018confocal}, FK~\cite{lindell2019wave}, and RSD~\cite{liu2020phasor}; and deep-learning-based methods: LFE~\cite{chen2020learned}, NeTF~\cite{shen2021non}, I-K~\cite{yu2023enhancing}. The implementations of the baseline methods follow their publicly available codes. LFE and I-K are trained on the same simulated data as ours, while NeTF is trained on the test measurement without extra training data. We only include NeTF for real-world experiments due to its computational burden for generating the results on hundreds of simulated scenes. NLOST is the previous version of this work, which was trained on data with the spatial resolution of 128$\times$128. The results are interpolated to 256$\times$256 for comparison.
By enlarging the support of the input size of the model, we further retrain NLOST~\cite{li2023nlost} on the new synthetic data with the spatial resolution of 256$\times$256, termed as NLOST*.

\setlength{\tabcolsep}{3.5pt}
\begin{table}[t]
  \begin{center}
  \small
  \caption{Ablation results on spatio-temporal attention. Spa and Tem denote that the attention mechanism only operates on the spatial or temporal dimension. }
  \label{table:validate_att}
  \begin{tabular}{cc|c|cc|cc}
  \hline
   \multicolumn{2}{c|}{Attention} & \multirow{2}{*}{Integration} &  \multicolumn{2}{c}{ Intensity} &  \multicolumn{2}{c}{ Depth}\\
   \cline{4-7}
    S.&T.& & PSNR$\uparrow$ & SSIM$\uparrow$ & RMSE$\downarrow$ & MAD$\downarrow$  \\
    \hline
    \checkmark & × & LGInt(Ours) & 24.01 &0.8529& 0.0877&0.0304\\
     × & \checkmark & LGInt(Ours) &\textbf{24.04} & 0.8499 &
    \textbf{0.0849}&0.0293 \\
    \checkmark&\checkmark& LGInt(Ours)  &24.03&\textbf{0.8583}&\textbf{0.0849}&\textbf{0.0292}\\
    \hline
    \checkmark&\checkmark& NoInt  &23.99 &0.8513 &0.0904&0.0365\\
    \checkmark&\checkmark& LocInt  &24.02 &0.8500 &0.0896&0.0361\\
    \checkmark & \checkmark & GloInt& \textbf{24.04}&0.8341&0.0882& 0.0324\\
    \checkmark&\checkmark& LGInt(Ours)  &24.03&\textbf{0.8583}&\textbf{0.0849}&\textbf{0.0292}\\
\hline
\end{tabular}
\end{center}
\end{table}


\subsubsection{Simulated Results}
We first evaluate our method on the \textbf{Seen} test data. The quantitative results of different methods are listed in Table~\ref{table:syn_all}. As can be seen, TRT-NLOS achieves the best performance (except for MAD), followed by NLOST*. Further, we also provide the quantitative results on the \textbf{Unseen} test data in Table~\ref{table:syn_all}, which includes more complicated scenes. As can be seen, TRT-NLOS performs the best in terms of all metrics for both intensity and depth reconstruction, demonstrating the superior generalization capability of the transformer architecture to unseen objects. For the intensity image, TRT-NLOS improves the reconstruction performance by a large margin over the physics-based methods, i.e., 2.75dB over FK and 3.23dB over RSD in terms of PSNR, which demonstrates the superiority of modeling the NLOS reconstruction with transformer. Meanwhile, compared with the deep-learning-based methods, TRT-NLOS achieves 1.83dB and 1.24dB improvements over I-K and NLOST*. For the depth map, TRT-NLOS
decreases RMSE by 13.89\% and 10.12\% over I-K and NLOST*, which demonstrates its effectiveness of exploiting local and global correlations in transient measurements.
 
In addition to the quantitative comparisons, we also provide qualitative results for reconstructed intensity images and depth maps, as shown in Fig.~\ref{fig:syn_all_int}. For the intensity image, LCT generate blurry results. FK and RSD recover main structures but without details. LFE outperforms physics-based methods but still fails to capture certain parts of the objects. I-K captures the primary structures of hidden objects, but the finer details are missing. In contrast, both NLOST* and TRT-NLOS deliver significantly improved results, with TRT-NLOS accurately recovering not only the overall object structures but also the fine-grained details. For the depth map, the traditional methods struggle to reconstruct the details, resulting in large error maps. The deep learning methods LFE and I-K mitigate this issue in most regions. NLOST* offers further improvements over these baselines. Notably, TRT-NLOS captures more comprehensive depth information and produces the most accurate and detailed reconstructions among all methods.

\subsubsection{Ablation Study} Our proposed TRT-NLOS outperforms NLOST* in most metrics (except for MAD), demonstrating the effectiveness of the newly designed transient measurement denoiser. Without the denoiser, we further conduct fine-grained ablation experiments to validate the efficiency of the spatio-temporal attention mechanisms, local-global feature integration, and shallow-deep feature fusion.

\setlength{\tabcolsep}{5.5pt}
\begin{table}[t]
\small
  \begin{center}
  \caption{Ablation results on shallow-deep feature fusion. Glo, Loc, and Sha indicate the deep global features, deep local features and shallow features, respectively.}
  \label{table:validate_block}
  \begin{tabular}{ccc|cc|cc}
  \hline
  \multicolumn{3}{c|}{Feature} & \multicolumn{2}{c}{Intensity}  & \multicolumn{2}{c}{Depth}\\
  \cline{4-7}
    Glo & Loc & Sha & PSNR$\uparrow$ &  SSIM$\uparrow$ & RMSE$\downarrow$ & MAD$\downarrow$  \\
    \hline
    \checkmark & ×& ×  & 24.00 &  0.8555 & 0.0871 & 0.0342\\
    \checkmark & \checkmark & ×  & 24.02 &  0.8553 & 0.0857 & 0.0315\\ 
    \checkmark & \checkmark&\checkmark&\textbf{24.03}&\textbf{0.8583}&\textbf{0.0849}&\textbf{0.0292} \\
  \hline
\end{tabular}
\end{center}
\end{table}

\noindent
\textbf{Spatio-Temporal Attention.} 
\label{abla_spa}The spatio-temporal self and cross attention mechanisms are designed to exploit the local and global correlations in 3D transient measurements in both spatial and temporal dimensions. We thus investigate the efficiency of individual spatial and temporal attention in the self-attention encoder and the cross attention decoder, with results listed in Table~\ref{table:validate_att}. As can be seen, spatial and temporal attentions contribute differently to the reconstruction performance, i.e., spatial attention improves intensity recovery, while temporal attention enhances depth estimation. Furthermore, combining both attentions results in a significant boost in overall performance.


\begin{figure*}[!t]
\centering
\includegraphics[width=0.98\textwidth]{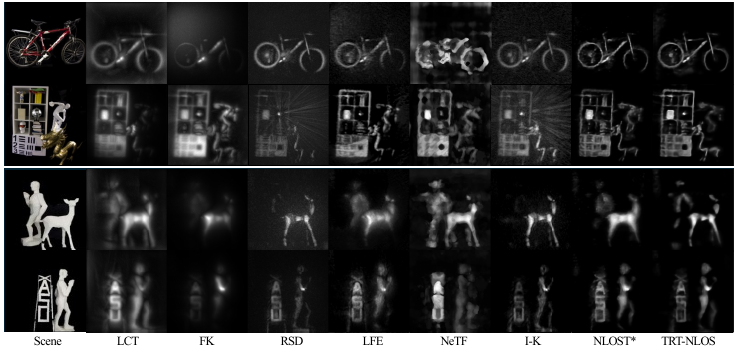}
\captionsetup{font={small}, justification=raggedright}
\caption{ Reconstructed hidden scenes from the public real-world data in~\cite{lindell2019wave} and the data captured by our NLOS imaging system. Zoom in for details.} 
\label{fig:real_results_fk}
\end{figure*}

\noindent
\textbf{Local-global Feature Integration.} In the cross attention decoder, the local and global features are integrated by querying each other in the token space. We further investigate the effectiveness of this integration (LGInt) by comparing it with other alternatives: NoInt (no integration between two branches), LocInt (only integration on local branch), and GloInt (only integration on global branch). The results are listed in Table~\ref{table:validate_att}. NoInt performs the worst in both intensity and depth, while the performance improves with one kind of information being integrated. When both local and global information is integrated, the performance is further promoted.
\label{abla_int}
\noindent
\textbf{Shallow-deep Feature Fusion.} We fuse shallow and deep features to recover the 3D volume of hidden objects. We thus study their contributions to the reconstruction performance in Table~\ref{table:validate_block}. As can be seen, the shallow and deep features contribute differently to the intensity and depth reconstructions, and fusing both of them further improves the performance, which demonstrates the efficiency of our shallow-deep fusion. 

\subsection{Experiments on Real-world Data}
\noindent
\textbf{Imaging System.}
\begin{figure}[!t]
\centering
\includegraphics[width=0.48\textwidth]{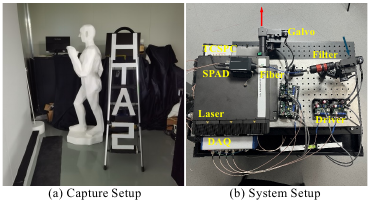}
\captionsetup{font={small}, justification=raggedright}
\caption{The capture setup and our self-built imaging system.} 
\label{fig:nlos_sys}
\end{figure}
For the evaluation on real-world data, we develop an active confocal NLOS imaging system. The prototype is illustrated in Fig.~\ref{fig:nlos_sys}. The system utilizes a 532 nm laser (VisUV-532) to generate pulses with a width of 85 picoseconds and a repetition frequency of 20 MHz, delivering an average power of 750 mW. These pulses are directed through a two-axis raster-scanning Galvo mirror (Thorlabs GVS212) towards the relay wall. Subsequently, both direct and indirect diffuse photons are gathered by another two-axis Galvo mirror, coupled into a multimode optical fiber, and then channelled into a SPAD detector (PD-100-CTE-FC) with a detection efficiency of approximately 45\%. The movement of both Galvo mirrors is synchronized and controlled by a National Instruments acquisition device (NI-DAQ USB-6343). The TCSPC (Time Tagger Ultra) captures the pixel trigger signals from DAQ, the synchronization signals from the laser, and photon detection signals from the SPAD. The temporal resolution of the overall system is approximately 95 ps. During data collection, the illuminated and sampling points maintain a consistent direction but are intentionally offset slightly to prevent interference from directly reflected photons during scanning. We perform a raster scan across a 256$\times$256 square grid of points on the relay wall. Each scanning point is allotted 1 ms for exposure, and the histogram is with a length of 512 bins and a bin width of 32 ps. We capture 6 different scenes with a self-built confocal imaging system, the data is released to facilitate future researches in this field.
To demonstrate the generalization capability of our method, We also use the public real-world dataset from Lindell~\etal~\cite{lindell2019wave}.

\noindent
\textbf{Real-world Results.}
The qualitative results on real-world data are shown in Fig.~\ref{fig:real_results_fk}. As can be seen, LCT generate the most scenes with blurry results. FK and RSD can reconstruct main structures but suffer from heavy noise. NeTF can only recover cursory shapes. LFE and I-K behaves better than the above methods but still misses some details. 
Both NLOST* and TRT-NLOS generate promising results, while the latter better recovers fine details and sharp boundaries of the hidden scenes, especially the girder of the bike, the bookshelf, and the pedestrian. The encouraging results produced by our method demonstrate its superiority over existing solutions.


\section{Conclusion and Future Work}

In this work, we introduce a novel time-resolved transformer (TRT) designed to enhance the performance of 3D reconstruction from transient measurements. By leveraging two intricate spatio-temporal designs, TRT effectively captures both local and global correlations within the data, addressing the challenges posed by low sensor quantum efficiency and high environmental noise. Based on TRT, we develop two mbodiments, TRT-LOS and TRT-NLOS, tailored for line-of-sight (LOS) and non-line-of-sight (NLOS) imaging tasks, respectively. 
The experimental results demonstrate that the proposed TRT-LOS and TRT-NLOS networks consistently outperform existing methods, showcasing their superiority in both accuracy and robustness for 3D reconstruction from transient measurements. This work paves the way for more accurate and efficient 3D reconstructions in challenging scenarios, particularly for long-distance and complex environments. Future researches could explore further optimization of the TRT, as well as its application to other domains requiring fine-grained spatial and temporal data integration.

\bibliographystyle{unsrt}
\bibliography{bare_jrnl_new_sample4}

\end{document}